\documentclass[pdflatex,sn-mathphys-num]{sn-jnl}

\usepackage{graphicx}%
\usepackage{multirow}%
\usepackage{amsmath,amssymb,amsfonts}%
\usepackage{amsthm}%
\usepackage{mathrsfs}%
\usepackage[title]{appendix}%
\usepackage{xcolor}%
\usepackage{textcomp}%
\usepackage{manyfoot}%
\usepackage{lmodern}%
\usepackage{anyfontsize}%
\usepackage{booktabs}%
\usepackage{algorithm}%
\usepackage{algorithmicx}%
\usepackage{algpseudocode}%
\usepackage{listings}%

\theoremstyle{plain}%

\theoremstyle{definition}%


\begin{document}

\title[Article Title]{Vectoring Languages}


\author*[1]{\fnm{Joseph} \sur{Chen}}\email{sorevan@g.ecc.u-tokyo.ac.jp}

\affil*[1]{\orgname{The University of Tokyo}}


\abstract{Recent breakthroughs in large language models (LLM) have stirred up global attention, and the research has been accelerating non-stop since then. Philosophers and psychologists have also been researching the structure of language for decades, but they are having a hard time finding a theory that directly benefits from the breakthroughs of LLMs. In this article, we propose a novel structure of language that reflects well on the mechanisms behind language models and go on to show that this structure is also better at capturing the diverse nature of language compared to previous methods. An analogy of linear algebra is adapted to strengthen the basis of this perspective. We further argue about the difference between this perspective and the design philosophy for current language models. Lastly, we discuss how this perspective can lead us to research directions that may accelerate the improvements of science fastest.}




\maketitle

\section{Introduction}\label{sec1}

Imagine that you are playing a word-guessing game with your friend. You chose the past tense of ``read" (the one that sounds similar to ``red") as your answer for her to guess.

``Is the answer close to `said'?", she asked. You said yes, since the two sound alike.

``Is the answer close to `real'?", she asked. You said yes since the two have similar spellings.

``Is the answer close to `look'?", she asked. You said yes since the two have similar meanings. 

``Is the answer close to `book'?", she asked. You said yes since the two often share appearances.

``Is the answer `read'?", she asked. You said yes.\newline

We use language every day so naturally that we sometimes do not notice how versatile we can recognize words. We can easily specify a word by comparing its similarity with other words, even if the comparisons we made are based on different reasons.\newline

The outline of this article will go as follows. First, I will start by introducing vectoring and its similar concepts. Then I will take a look at related works and the concerns they propose. In the third section, I will construct our vectoring view of language by starting with some taxonomy and definitions, then proceed to show how a vectoring view can be a good approach to understanding language. Lastly, I will conclude with some inspirations we gain from vectoring, and present the path to a few potential future works.\newline

Philosophers have been long discussing about the meaning and usage of specific words and how they resemble the talker's internal stance. For instance, how the truth value is affected by the word ``but" in a sentence, or when can we say ``so-and-so means so-and-so". I want to argue that the natural way humans use language is by considering multiple attributes of language embedded in its high-dimension vector property. Either consciously or subconsciously, we tend to consider how fluent and natural our sentence is when speaking or writing, how the words sound when composing lyrics, and for university students, how long each word or sentence is when writing semester reports.  Recently, using vectors as word representation has empirically achieved great results in the form of large language models (LLMs). This outcome hints that the idea backing up what AI scientists have been working on may be much closer to the essence of language than what we expected. For the sake of simplicity, I will call the idea of treating words as high-dimensional vectors, and a language as a high-dimensional vector space, ``vectoring".

To make our metaphor with the vector space more concrete and to strengthen the connection between our philosophical research and the scientific basis of AI science, we will be loosely adapting some definitions from linear algebra. I believe the amount of linear algebra introduced in this article will not be enough to overwhelm readers without the corresponding background, but enough for discovering new aspects of language that were previously unnoticed.\newline

\section{Related Works}\label{sec2}
\subsection{Word vector representations as a theory}\label{sub1}
Due to how our work crosses different fields of interest, prior research on the philosophy of language, philosophy of mind, natural language processing(NLP), large language models(LLM), linear algebra, and neural science brought us significant reference materials. It is also worth mentioning that during our research on related approaches, we also came across several distinguished articles attempting to bridge the gap between the language space of AI models and human philosophers. \newline

This work is greatly influenced by ``Plato's Camera: How the Physical Brain Captures a Landscape of Abstract Universals" by Paul M. Churchland \cite{Churchland2013-CHUPCH-2}. In his work, Churchland has distinctly pointed out a possibility of how the neural system in the human body ``manage to generate a ‘language space,’". However, Churchland's publication stayed more general and connected to neuroscience and neural network structures, which leads to problems generated by how differently humans and machines perceive the world\cite{oai:repository.dl.itc.u-tokyo.ac.jp:02003243}. I want to focus on specifically the language vector space and its connection to AI language models in this article. Churchland also uses the word \textit{map} to represent a ``high-dimensional structural homomorph of the objective similarity-and-difference structure of some objective, abstract feature-space", where I instead want to use the terms ``Function" or ``Target function" for a more consistent taxonomy, since maps in computer science often refer to an injection relationship rather than geographical landscapes.\newline

Goodman et al. (2014) \cite{Goodman2014-wu} originated the connection between the computational-based language and the language structure in the human mind. Brenden M. Lake and Gregory L. Murphy (2023)\cite{Lake2023-fw} specifically pointed out that psychologists showed great interest in contemporary NLP systems serving as psychological theories, and that while they are fairly successful models of human word similarity, they fall short in many other respects. Kyle Mahowald et al.(2024) \cite{Mahowald2024-zf} also 
compared the linguistic abilities of AI models and human minds and grounded the distinction of the two to neural science. These works share a common suffering in that it is difficult to compare the human understanding of language and the implementation of contemporary AI language models, due to the lack of a common language structure that gives us an apples-to-apples comparison. 

Wong et al.(2023) \cite{Wong2023-bp} proposed a rational meaning computational framework for language-informed thinking that combines neural language models with probabilistic models for rational inference, using the probabilistic language of thought (PLoT) framework introduced in \cite{Goodman2014-wu}, which models thinking as probabilistic programs rather than high dimensional model spaces. Brandt(2018)\cite{Brandt2018-gg} also proposes a novel language structure based on Culioli's ``Theory of Enunciative Operations", but focuses more on the syntactic relationship of words in sentences.

\subsection{Word representations in Machines}\label{sub2}
How machines understand natural language is one of the most active fields in AI science, centering around how AI can communicate naturally with humans, understand humans' thoughts, and responding their output in languages that humans can fully understand \cite{Lake2017-wr}\cite{Krafft2016-jb}. These can include the forms of videos, images, audio, and understanding of environments, but mostly we want to focus on text language.
\cite{Schwettmann2018-in}\cite{Creswell2022-vh}\cite{Caliskan2017-gy}\cite{Caucheteux2022-dt}\cite{BrownPeter1993-on}\cite{Lake2023-fw}
We do want to point out that our proposed structure is not limited to text language, and we expect future works to be augmented with other aspects of natural language. 

\subsection{The danger of LLMs}\label{sub3}
A significant number of research papers we came across address the danger of  LLMs, most of which originate from the distinction of language understanding between humans and AIs\cite{Gehman2020-tw}\cite{Lazaridou2021-vy}\cite{Weidinger2022-ig}\cite{Bender2021-gr}. The work, ``Word Meaning in Minds and Machines" by Lake et al.(2023)\cite{Lake2023-fw} specifically pointed out some flaws in the word vector representation of AI language models, and we will address these flaws in later chapters in this article.

\section{Vectoring}\label{sec3}
Using vectors as a representation of words has a long history, dating back to research in computer science in the 1980s to the 1990s\cite{Hinton1986-eb}\cite{Elman1990-lu}. However, language in philosophy still remains anthropocentric, and benefits little from the breakthroughs of computer science research about language models. In this section, we will propose our approach of a vectoring perspective on languages, and we will show how this approach directly connects to recent AI science breakthroughs in later sections.

There are many aspects of language like the meaning of a word, the sound that happens when we utter a word, the aesthetic of poetry, and so on. Not only that we cannot \textit{consider} all aspects of language at once, but we, as mortals, also cannot \textit{recognize} all aspects of language. We are like blindfolded people trying to figure out what an elephant looks like just by touching it. Since this symbol is a high dimensional space with an unknown shape that changes dramatically depending on the perspective we look at it, we shall conveniently call it the vector space of a language, or $V_L$ for short.\footnote{
We say that a language is represented as a vector space $V_L$.  I believe it is not that relevant to formally define the vector space in a mathematics fashion, since it is still difficult for us to understand what elements in this vector space stand for. I suggest that this vector space is over a scalar field $F$ representing the times of occurrences and that the addition operation stands for the idea when combining vectors without considering the sequential pattern, but won't formally cover it in this article. This part will be left as future work for others.} \newline

An important concept of vectoring is the projection. Since we are not able to observe the $V_L$ space as a whole, we project $V_L$ into a subspace with a potentially lower dimension in order to understand it better. We can think of casting a shadow on the wall in order to better understand the shape of an object. An example of a projection of $V_L$ is to focus on the meaning of words.
In linear algebra, a projection is a linear transformation $P$ in which applying the projection maps vectors in $V_L$ into a subspace $W$. Additionally, applying the projection twice leaves its image unchanged compared to applying it once. These harmonies with our analogy of the projection: trying to find the meaning of the meaning subspace itself essentially brings us the same subspace. \footnote{In linear algebra, a projection $P$ is defined as a linear operator on $V$ such that $P: V \rightarrow V$ and that $P^2 = P$. This means that the projection on $V_L$ *can* maintain its rank so that the subspace that projection $P$ maps to is the original space, nothing less.}\newline

Words, being the basic element in nearly all other research programs about similar topics, are just a set (since we are using a plural form of word) of vectors, and aren't much different from utterances or phrases, which are also some set of vectors in $V_L$. However, a word is a good example of what a single vector may represent, and we will talk more about it later in (link to later section). For now, we will take a little detour to how vectoring is currently used in AI science, which is what we will call \textit{practical vectoring} in this article.

\section{Practical vectoring}\label{sec4}
The development of language models in computer science has come a long way, and by the underlying method behind the contemporary models, we can roughly divide the technology history into two stages. We will, however, later show that the two methods are essentially the same.

\subsection{Word2Vec}\label{subsec2}
Arguably one of the most important breakthroughs in language models is the work by Mikolov et al. in 2013 \cite{Mikolov2013-rg}. This work exceeds the data size and training efficiency of previous works in machine-learning language models by orders of magnitudes.  It is shown that words can have \textit{multiple degrees of similarity} (which origins from the authors' previous work, \cite{Mikolov2013-lh}).\newline

The resulting model, named Word2Vec, achieved great success in preserving the linear regularities among words. This can be shown when performing arithmetic operations on word embedding vectors, we get results that somehow reflect a good understanding of the meaning of the words.  

For example, vector(``King") - vector(``Man") + vector(``Woman") results in a vector that is closest to the vector representation of the word Queen. This method served as a paradigm for language models for about three years until transformers were introduced and text-generative AI experienced another big breakthrough.

\subsection{Transformers}\label{subsec3}

The idea behind text generation AI models' architecture is to repeatedly predict the token that should appear next. A ``token" mentioned in AI language models is a sub-word or component that makes up a word. For example, the word ``unnerving" can be divided into three tokens: ``un", ``nerv", and ``ing", each serving its own purpose in constructing the word. The main structure in these LLMs is called Transformers, and Transformers are used to capture the context (often considered by scientists as a few words before where the prediction takes place. However, in reality, words coming from early in the input sequence can also have an impact in late predictions) for the next prediction until a stop signal is generated. The output of the Transformer is a high dimensional vector, for our reference, GPT-3\cite{Brown2020-th} by openAI uses a vector length of 12,288 to represent each token. \newline

In plain sight, auto-regressive models that are trained to take in previous context and predict the next token seems different from what Word2Vec did: capturing the relationship and similarities between words. We can see in practice that LLMs do seem to encode correlated tokens in close locations, so there has to be a connection between the two. In fact, there is a strong connection between the two.\newline

This connection originates from the ``self-attention block", which is the most important component in the Transformation structure. The self-attention block has a mechanism for learning a new token embedding space by linearly recombining token embeddings from some prior space. The weights that apply to the linear combination give higher importance to the tokens that are originally close together in that space. This results in pulling the vectors that have closer distances further together and the token correlation relationship will be transformed into embedding proximity relationships. The learning process of Transformers includes learning on a series of incrementally refined embedding spaces, where each space is based on reassembling the elements from the previous one.\cite{Vaswani2017-tf} \cite{Chollet2023-hd}

There are two important properties that the self-attention block brings to the Transformer structure: continuity and linear interpolativity. This means that the embedding space that the Transformers will ultimately get is \textit{semantically continuous} and \textit{semantically interpolative}. In other words, moving a bit in embedding spaces only changes the meaning of the corresponding tokens by a bit, and the linear combination of two vectors yields a vector corresponding to a token with roughly the same meaning as the linear combination of the two meanings corresponding to the two original vectors. The Word2Vec space also verifies for these properties.

\subsection{But how does a machine learning model learn meanings from data?}\label{subsec4}
The process of learning from training data is a common practice among all the AI applications invented today. But this does not stop us from wondering why this also works on word representations.
How exactly do the models learn from, and \textit{why} can they learn the meaning from these data? Computer scientists did not manually label all possible relationships between all possible words for AI models to learn from. Instead, they feed the learning process lots of existing documents. This means that what a language model actually learns, is the \textit{pragmatic appearance distribution of each word}.
Even though we are still not sure how the pragmatic appearance yields meaning, there is a hypothesis called the distributional hypothesis that we can refer to (Gastaldi JL, 2021)\cite{Gastaldi2021-sy}. According to the distributional hypothesis, linguistic items with similar distributions can yield similar meanings. If we adopt a distributionalist's view here, we can believe that there exists a positive correlation between the probabilistic distribution of the coming word and the actual meaning of the word.

\section{Differences between vectoring and practical vectoring}\label{sec5}
It is a nice timing to explain here a popular misunderstanding: the difference between vectoring and practical vectoring. While using similar wording and wanting to achieve similar goals, there are still large differences that shouldn't be ignored when discussing vectoring. \newline

\textbf{Representation dimensions}
In vectoring, a word is represented as a high dimensional vector, which captures \textit{all degrees} of a word in the language vector space. In practical vectoring, however, humans defined the dimension. The standard Word2Vec pre-trained models have 300 dimensions\cite{Pennington2014-lv}, and the optimal dimension count is experienced to be dependent on the quality and quantity of the training data.\newline

\textbf{Empirical data}
Currently, the majority of the training data utilized is in the form of text data, especially those written on the internet. The ease of data access leads to a distribution of language usage that differs from real-world usages, which may lead to inaccurate modeling of the language space $V_L$. For example, current large language models generally achieve higher accuracy in written language than in spoken language.\newline

\textbf{Structural restrictions}
Another difference between practical vectoring and vectoring is the structural restrictions that inevitably exist when implementing computer language models. It doesn't matter whether you choose a Skip Gram structure for implementing Word2Vec or a Transformer structure for implementing large language models, the model structures all have fixed dimensions, and use floating point numbers to represent the vectors. What's more important, is that the hidden layers of either structure hardly mean anything to a human, while in vectoring, we can explicitly define our projections for any kind, such as ``the meaning of a word used in France in 1750". \newline

We will show how the differences make us free from the problems that are illustrated in Brenden M. Lake, and Gregory L. Murphy, 2023\cite{Lake2023-fw}.

\section{Taxonomy and Definition}\label{sec6}
\begin{enumerate}
    \item The \textbf{vector space of a language}, or $V_L$ in this article, is a comprehensive model of a language $L$ in our interest. Due to its high dimensional nature, humans are not able to observe the vector space. Instead, we are \textbf{projecting} the vector space into a subspace with a lower dimension, and argue reasonably within the subspace.
    \item A \textbf{Projection} is a linear transformation $P$ where $P: V_L \rightarrow W$, in which $W$ denotes a subspace of $V_L$ that has a dimension not larger than $V_L$. \footnote{In linear algebra, two projections on the same space with the same rank grant two isomorphic subspaces.}
    \item When we say we are interested in an attribute of $V_L$, for example, the meaning of words, we are finding a subspace $W$ of $V_L$ where there exists a function that takes in a word as an input and gives a set of vectors as an output. Formally speaking, we are looking at an attribute subspace $W_a$ such that there exists a function $F(w) \rightarrow \{\{y\} | y \in W_a\}$.
    \item A \textbf{Word} $w$ is a single vector in $V_L$. 

\end{enumerate}
Given the above definitions, we can see that:
\begin{enumerate}
    \item We can denote the ``finding of all meanings of a word" can be expressed as $F(w) \rightarrow \{\{y\} | y \in W_{meaning}\}$, whereas ``finding the most common meaning of a word" can be expressed as $F(w) \rightarrow \{y | y \in W_{meaning}\}$.
    \item Existing theories are just a further projection of the attribute subspace $W_a$ when we find that the dimension of $W_a$ still poses difficulties for us to understand. Note that a projection of $W_a$ also means that it is a valid projection of $V_L$.
    \item The way we can perform a projection is by the explication of the taxonomy and definition. It needs to include what we are interested in, what we are not, and a clear perspective on how we should consider this attribute. (For example)
\end{enumerate}

By adopting a vectoring view of language, we can see that it is easy to address some hard logical problems that other perspectives suffer from. For example, we are historically concerned about the truth value of the sentence ``The king of France is wise". On the one hand, we can say that when projecting to Russell's theory of referring, we get a result that the sentence is nonsense. On the other hand, we can directly say that we are projecting this sentence to a subspace of a function that gives us the truth value of the input, formally: $F_{truth}(x) \rightarrow {True, False}$. Then we can simply regard the sentence as neither True nor False but not well-defined for the function since the subject ``The king of France" is undefined.

\section{Response to Word Meaning in Minds and Machines}\label{sec13}
In the work ``Word Meaning in Minds and Machines" by Lake et al.(2023)\cite{Lake2023-fw}, the author proposes five reasons why the word representations generated from large language models do not represent word meanings. I will argue that while they do indeed exist in practical vectoring (LLMs), it is not that significant when we take vectoring as a perspective. The five reasons are:

\subsection{Word Representations Should Support Describing a Perceptually Present Scenario or Understanding Such a Description}\label{ss1}
Word representations of practical vectoring were criticized for the lack of its interactive interface which can be used to gather information from the environment, rather than only depending on text input sequences. This problem is also addressed in our discussion about the differences between vectoring and practical vectoring, where biased empirical data can lead to biased distribution in the vectors of word representation. In vectoring, which we discussed earlier, however, there exist subspaces that address scenario understanding.

\subsection{Word Representations Should Support Choosing Words on the Basis of Internal Desires, Goals, or Plans}\label{ss2}
Assuming that folk psychology holds, the word choices we make should somehow reflect how we think internally. This is not the case in how practical vectoring works, as it generates text based on a calculated probability distribution over tokens. There is still space for arguments about whether there is space for internal states in the vectoring perspective, but we can easily define a function $F(s,t)$ such that $F$ denotes how well a sentence $s$ can express to someone that the speaker wanted the task $t$ to be finished.

\subsection{Word Representations Should Support Responding to Instructions Appropriately}\label{ss3}
Current large language models fail to transform text into real actions, which can be seen as a piece of evidence that language models couldn't really learn the meaning behind what they generated. However, it is arguable if actions responding to instructions do have direct causal relationships with the validity of a word representation method. Imagine a great mathematician sitting at his desk solving problems no one has solved. Although he does answer your questions thrown at him, he doesn't respond to you asking him to leave his seat so you can clean the empty energy drink cans under his desk. In this case, we still say that his words are meaningful, even if he ignores to perform any action you asked him to do.

\subsection{Word Representations Should Support Producing and Understanding Novel Conceptual Combinations}\label{ss4}
The authors argue that large language models can not generate new words other than those they are already programmed to (in GPT3's case that is the 12288 possible tokens). This is true since the dimension count is defined by humans, and it is not in the nature of current AI models to generate novel outcomes not existing in the training data. This, too, doesn't apply to our vectoring perspective, as we leave it to the nature of the language to define the dimensions itself.

\subsection{Word Representations Should Support Changing One’s Beliefs About the World Based on Linguistic Input}\label{ss5}
The last point addresses a difference between vectoring and practical vectoring that we didn't cover in previous sections. A fully trained language model will *not* change its structure or weight when interacting with others. Although we can consider past conversations by using them as context for the model to consider, the weights (a.k.a. the learning outcome of the model) will not change. The only way around this constraint is to design an AI agent system that retrains the network after new interactions. Unfortunately, this is currently too costly both in time and resources, and it is unlikely to be realized. This point also shows that language is not static, and its components are forever changing. We are very excited about how AI scientists will take on this problem, and the future research in this field.

\section{Conclusion}\label{sec14}
The vectoring view of language showed great empirical achievements through the success of recent large language models. We can see that regardless of the different definitions our current theories have for word meanings, large language models can give us the meanings we want. This opens up the possibility that there are some definitions of attributes of language lying above what we can capture using language theories but can be approximated using practical vectoring.

The vectoring view of language also tells us crucial clues on how we can proceed with our research of the language. Theories that focus on similar topics generate projections with high correlation. For example, the theory of meaning and the theory of referring. While these theories discuss important topics of language, newer theories tend to give less contribution to our overall understanding of language given the high correlation with previous research. Alternatively, if we can combine multiple perspectives from totally different approaches, we can better understand the hidden shape of language thanks to the low correlation. What’s more, by tying the philosophy of language together with the rapidly innovating large language models, we not only provide a sturdy theory basis for AI but also gain access to unlimited experiments on how the language works, in the form of training new models with novel assumptions of the vector space of language.

\backmatter

\bibliography{sn-bibliography}

\end{document}